\documentclass{article}
\ifdefined\pdfsuppressptexinfo \pdfsuppressptexinfo=-1 \fi
\usepackage{iclr2027_conference,times}

\usepackage{microtype}
\usepackage{graphicx}
\usepackage{subcaption}
\usepackage{booktabs}
\usepackage{multirow}
\usepackage{array}
\usepackage{bm}
\usepackage{amsmath,amssymb,amsthm,mathtools}
\usepackage{algorithm}
\usepackage{algorithmic}
\usepackage{xcolor}
\usepackage{makecell}
\usepackage{bbm}
\PassOptionsToPackage{hyphens}{url}\usepackage{url}
\usepackage{hyperref}

\newcommand{\x}{\mathbf{x}}

\newcommand{\thetab}{\boldsymbol{\theta}}
\newcommand{\phib}{\boldsymbol{\phi}}

\newcommand{\argmax}{\operatorname*{argmax}}
\newcommand{\rt}{\mathrm{RT}}
\newcommand{\E}{\mathbb{E}}

\newcommand{\corr}{\operatorname{corr}}
\newcommand{\flip}{\varsigma}
\newcommand{\Fspace}{\mathcal{F}}          %

\title{Recovery-Directed Symbolic Distillation\\of Neural Likelihoods}

\author{Kiant\'e Fernandez\thanks{Corresponding author.} \\
Department of Psychology \\
University of California, Los Angeles \\
Los Angeles, CA 90095, USA \\
\texttt{kiante@ucla.edu}
\And
Xinwei Li \\
Department of Civil and Environmental Engineering \\
National University of Singapore \\
Singapore \\
\texttt{xinwei.li@u.nus.edu}}

\iclrfinalcopy %

\begin{document}

\maketitle
\lhead{Preprint}

\begin{abstract}
Amortized neural likelihoods enable computationally expensive inference for models with analytically intractable or unspecified likelihoods, but their black-box nature limits interpretability. We introduce a symbolic distillation pipeline that converts trained neural likelihoods into explicit, interpretable expressions optimized for efficient parameter estimation. Our approach uses a recovery-directed objective to guide symbolic regression toward expressions that preserve parameter-recovery accuracy rather than merely approximating the likelihood function. Candidate expressions are evaluated on held-out datasets and selected using a criterion that jointly accounts for expression complexity, parameter-recovery performance, and distributional distance from the learned likelihood. We evaluate the pipeline on the diffusion decision model, a classical cognitive model, whose analytically tractable likelihood provides ground truth for controlled evaluation. The proposed recovery-directed objective improves parameter recovery over standard symbolic-regression objectives. The resulting symbolic likelihoods enable over 100 times faster parameter evaluation than both neural likelihoods and, when available, the exact likelihood, while maintaining a manageable loss in precision. We further demonstrate these computational benefits in Bayesian hierarchical inference on empirical data. Our pipeline provides a lightweight interface for integrating symbolic distillation with existing neural-likelihood estimation methods and can be adapted to a range of simulation-based inference settings.
\end{abstract}

\section{Introduction}
\label{sec:intro}

Cognitive science formalizes hypotheses about mental processes as mechanistic models \citep{kriegeskorte2018cognitive}. Such models specify an explicit generative process $\bm{\mathcal{G}}(\bm{\theta},\bm{\epsilon})$ ($\bm{\epsilon}$ for stochastic noise) for observed behavior $\bm{y}$, encoding substantive hypotheses of underlying operations and indexed by a small set of interpretable parameters $\bm{\theta}$. To further validate the hypotheses, the models are often fit to experimental data, which requires estimating the parameters from observed data, typically through the likelihood $p(\bm{y}|\bm{\theta})$. However, many cognitive models with richer and more realistic assumptions fail to have a tractable likelihood.

One popular class of cognitive models, sequential sampling models (SSMs), demonstrates this tension. SSMs posit that decisions arise from the noisy accumulation of evidence over time until reaching a decision threshold \citep{ratcliff1978theory, forstmann2016sequential}. Under a forced-choice task with options $\mathcal{C}$, each observation $\bm{y} = (\rt, c)$ therefore comprises a response time 
$\rt$ and a chosen option $c\in \mathcal{C}$. SSMs are widely adopted across domains, including perceptual decision-making and clinical research \citep{ratcliff2016diffusion}, value-based and consumer choices  \citep[for reviews, see][]{clithero2018response,krajbich2026decomposing}, and moral dilemmas \citep[e.g.,][]{maksimenko2025video}. Nonetheless, only the simplest case of SSMs is analytically tractable in likelihood  $p(\rt,c|\bm{\theta})$, such as the Navarro--Fuss series \citep{navarro2009fast} for the classic drift-diffusion model \citep[DDM,][]{ratcliff2008diffusion}, whereas its scientifically interesting extensions (e.g., collapsing decision bounds and leaky accumulators) generally lack a closed-form likelihood.

Simulation-based inference (SBI) addresses the parameter estimation of cognitive models with infeasible likelihoods by using generative processes $\bm{\mathcal{G}}$ \citep[a review,][]{cranmer2020frontier}. The amortized variants (e.g., likelihood approximation networks \citep[LANs;][]{fengler2021lans}, neural density
estimators \citep{boelts2022flexible}, and amortized Bayesian workflows \citep{radev2020bayesflow, radev2023bayesflow}) have become particularly popular, owing to their reusability across datasets. That is, a neural network is trained only once on simulated data and can then be applied to any new dataset at negligible cost, without further simulation or retraining. 
Moreover, toolboxes like HSSM \citep{fengler2026hssm} for SSMs have integrated pretrained likelihood approximations $p_{\bm{\phi}}(\bm{y}|\bm{\theta})$ by amortized SBI into a
standard probabilistic-programming interface. However, the learned likelihood $p_{\bm{\phi}}(\bm{y}|\bm{\theta})$ is a black box, because the network weights 
 $\bm{\phi}$ carry no substantive interpretation. Consequently, the recovery analysis,
approximation diagnosis, and structural insight into the likelihood must all be done numerically.

Symbolic regression (SR) may offer a way to open this black box. SR is a recently growing supervised learning approach that discovers symbolic mathematical expressions $f$ from data $\left\{\big(\bm{x}, f_0(\bm{x})\big)\right\}$ by searching the space of closed-form expressions $\mathcal{F}$, trading off accuracy against expression complexity \citep[a review,][]{makke2024interpretable}. The resulting analytic alternatives $f\in \mathcal{F}^*$ are not only inherently interpretable and auditable but also differentiable and cheap to calculate. Whatever the search technique, the canonical objective  $\mathcal{L}(f,f_0)$ is to minimize the expected error between the candidate expression $f$ and the real value $f_0$. Although, from the perspective of SR, the probability density $p$ is simply another target function (i.e., $f_0(\bm{x}) = \log p (\bm{y}|\bm{\theta})$ and $ \bm{x} = (\bm{\theta},\bm{y})$), the standard objective $\mathcal{L}$ is misaligned with the purpose of obtaining closed-form approximations for the implicit cognitive models, namely efficient and accurate parameter estimation with an interpretable structure.

In this work, we propose a pipeline that distills analytical expressions from neural
approximations to make parameter estimation easier (\autoref{sec:method}). 
First, the neural likelihood $p_{\bm{\phi}}$ is admitted only after a sequence of SBI diagnostic checks. Then the symbolic regression search is guided by a recovery-directed loss function $\mathcal{L}_{\lambda}^r(f,\log p_{\bm{\phi}})$ (\autoref{sec:method-loss}), which substantially improves parameter recovery over the standard symbolic-regression objective (\autoref{sec:results-objective}). Finally, each candidate expression on the Pareto front $\mathcal{F}^*$ is fit with a full Bayesian sampler and ranked by how well it recovers the parameters on held-out data. We evaluate it on the simplest DDM, whose exact likelihood permits direct assessment of approximation accuracy and helps identify the sources of estimation error. The resulting symbolic likelihoods are tens of times faster to evaluate than compiled exact and neural likelihoods at the size of one subject, and more than 100 times faster on large datasets, while maintaining a manageable loss in precision (\autoref{sec:results-speed}).

\section{Method}
\label{sec:method}
\begin{figure}[bpth]
    \centering
    \includegraphics[width=\linewidth,
    trim={0.2cm 1cm 1cm 1cm},
  clip]{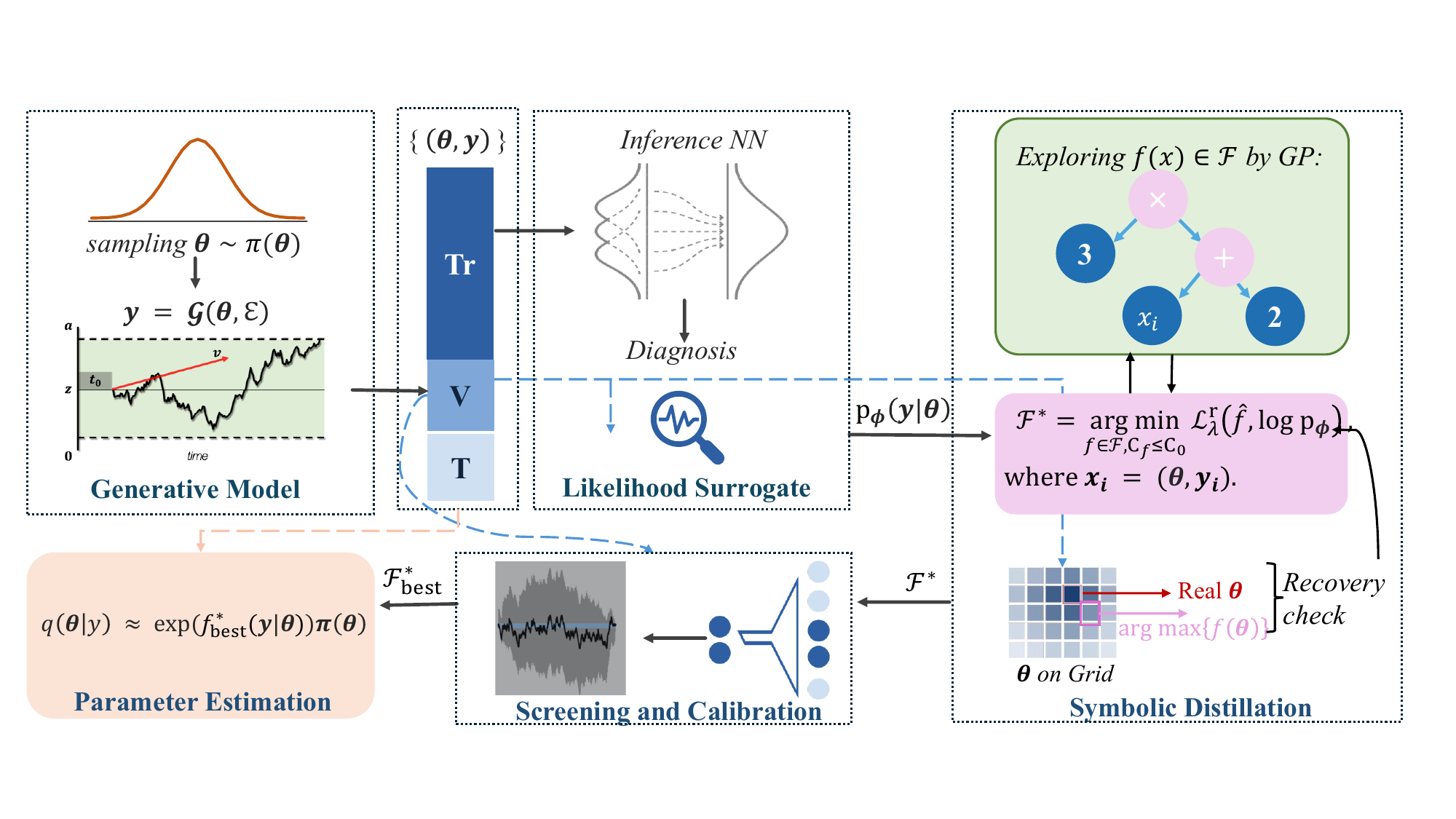}
    \caption{\textbf{Pipeline for recovery-directed symbolic distillation of the likelihood of a
generative model. } Tr, V and T denote the training, validation and test data. }
    \label{fig:pipline}
\end{figure}
Figure~\ref{fig:pipline} summarizes the pipeline, in which each stage consumes the output of the previous one. In Stage~1, a neural likelihood $p_{\bm{\phi}}(\bm{y}|\bm{\theta})$ is trained and checked. In Stage~2, symbolic regression under the recovery-directed loss $\mathcal{L}^r_{\lambda}$ produces candidate expressions $\mathcal{F}^*$. In Stages~3--4, each candidate is fit with the No-U-Turn Sampler \citep[NUTS;][]{hoffman2014nuts} in PyMC \citep{abrilpla2023pymc} and ranked by how well it recovers the parameters.

To set up, we place independent uniform priors $\bm{\pi}(\bm{\theta})$ on each parameter over its range $\Theta$. All data are simulated from $\bm{\mathcal{G}}$ in the group format $\left\{(\bm{\theta}^{(g)},\bm{y}_{1:K_{max}}^{(g)})\right\}_{1:G_{all}}$, in which each
parameter set $\bm{\theta}^{(g)} \sim \bm{\pi}(\bm{\theta})$ generates
$K_{max} = 1{,}000$ independent trials. Different stages use different groups, and the
number of trials drawn from each group depends on the stage (\autoref{app:data_split}).

\subsection{Stage 1: amortized SBI admitted by diagnostics}
\label{sec:method-nle}
\paragraph{Simulator Specification.} DDM \citep{ratcliff1978theory, ratcliff2008diffusion} describes the forced binary choice process, hence $\bm{y} = (\rt, c)$, where the choice outcome set $\mathcal{C} = \{-1, +1\}$. The DDM accumulates evidence at time $\tau$ by
\begin{equation}
d V(\tau) = v\,d\tau + d W(\tau), \qquad V(0) = a\,(2z - 1),
\end{equation}
where the within drift noise $W(\tau)$ is a Wiener process. The decision is determined when $|V|$ hits a boundary at $\pm a$. That is,
\begin{equation}
    \bm{\mathcal{G}}(\bm{\theta},\bm{\epsilon}) = \Bigg(\rt = t + \inf\big\{\tau : |V(\tau)| \ge a\big\}, c = 2\mathbbm{1}\big(V(\rt-t)>0\big)-1\Bigg).
\end{equation}
$\thetab = (v, a, z, t)$ contains drift rate $v$,
boundary half-separation $a$, relative
start point $z$, and non-decision time $t$. Notably, given the symmetric property of first passage time on both boundaries, the parameter $\theta$ can be identified with single-side joint density (i.e., $p(\rt,c = 1|\theta)$). For detailed settings, see \autoref{app:DDM}. 

\paragraph{Amortized SBI Structure.} Because $\bm{y} = (\rt, c)$ combines a continuous
and a discrete variable, the joint density factorizes exactly
\citep{boelts2022flexible} as
\begin{equation}
\label{eq:factorization}
\log p_{\phib}(\rt, c \mid \thetab) \;=\;
\underbrace{\log p_{\phib_\mathrm{flow}}(\rt \mid \thetab, c)}_{\text{normalizing flow}} \;+\;
\underbrace{\log P_{\phib_\mathrm{class}}(c \mid \thetab)}_{\text{classifier}}.
\end{equation}
We model the continuous part with a normalizing flow trained by conditional flow
matching \citep{dinh2017realnvp, papamakarios2021normalizing} and the discrete part with
a small classifier (architectures in \autoref{app:nle}). This module can be replaced by
any amortized SBI method that allows likelihood evaluation, including methods not
trained specifically for likelihood estimation, such as all-in-one SBI
\citep{gloeckler2024allinone}.

\paragraph{Surrogate Diagnosis.} So that approximation errors do not propagate to later
stages, the neural likelihood must pass a diagnostic gate on held-out validation data,
which checks the calibration of the RT flow and the choice classifier and how well the
parameters $\bm{\theta}$ are recovered (\autoref{app:nle}).

\subsection{Stage 2: Symbolic distillation for log-density}
\label{sec:method-search}
SR freely searches the space $\mathcal{F}$ of analytic expressions composed from a fixed operator set $\mathcal{O}$ and inputs $\bm{x}$, returning a Pareto front $\Fspace^*\subset \Fspace$ of candidates that trade off accuracy against complexity. For density approximation from a previous neural surrogate, the input data is $\bm{x}_i= (\bm{\theta}_i, \bm{y}_i)$, while the output is the log neural likelihood $\log p_{\phi}(\rt_i, c{=}1|\bm{\theta}_i)\in \mathbb{R}$. 

\paragraph{SR searching techniques.}  We use PySR~\citep{cranmer2023pysr}, which represents each candidate as an expression tree ($\bm{x}$ as leaves and $\mathcal{O}$ as nodes) and evolves several populations in parallel by genetic programming. The operator set $\mathcal{O}$ includes binary operators
$\{+, -, \times, \div\}$ and unary operators $\{\exp, \log, \sqrt{\cdot}\,, (\cdot)^2\}$. Runs are budgeted by the same time budget, and their Pareto
fronts are unioned and de-duplicated on the printed form denoted by $\mathcal{F}^*$. 

\paragraph{The recovery-directed objective $\mathcal{L}^r_{\lambda}(f,\log p_{\phib})$.} \label{sec:method-loss} Compared to the classic mean square error (MSE) loss function 
\begin{equation} \label{eq: loss_mse}
    \mathcal{L}(f,\log p_{\phib}) = \frac{1}{N}\sum_{i}\left(f(\bm{x_i})-\log  p_{\phi}(y_i|\bm{\theta}_i)\right)^2,
\end{equation}
the recovery-directed objective $\mathcal{L}^r_{\lambda}(f,\log p_{\bm{\phi}})$ in \autoref{eq:loss} adds two penalty terms, a recovery term $1-\mathcal{R}(f)$ weighted by the hyperparameter $\lambda$ and a constraint term $B(f)$,
\begin{equation}
\label{eq:loss}
\mathcal{L}^r_\lambda(f,\log  p_{\bm{\phi}}) = \underbrace{\tfrac{1}{N}\textstyle\sum_i \bigl(f(\bm{x_i}) - \log  p_{\phi}(y_i|\bm{\theta}_i)\bigr)^2}_{\text{density fit}}
\;+\; \lambda \bigl(1 - \mathcal{R}(f)\bigr)
\;+\; 10^{3} B(f),
\end{equation} 
where the last term enforces two hard constraints, with $B(f)$ counting violations.
First, every element of $\bm{x}_i$ must appear in $f$ after simplification (e.g., $(-h)+h=0$ removes $h$), since a parameter cannot be recovered from an expression that omits it. 
Second, $f$ may not contain constant leaves. Adding a constant to a log-likelihood does not change its estimates, so forbidding constants removes redundant expressions and the cost of optimizing them in PySR. Constant factors can still arise through simplification, as in $h+h=2h$.

$1-\mathcal{R}(f)$ stands for the approximated parameter recovery error given the grid recovery proxy on training data of the current candidate $f$. The proxy $\tilde{\theta}_{j}^{(g)}$ for the parameter set $\theta^{(g)}$ on $j_{th}$ element is derived in Equation~\ref{eq:profile} such that the summed log-likelihood under $f$ over $K$ training trials, $\sum_k f(\bm{y}_{k}^{(g)}; \{\theta_{-j}^{(g)},\tilde{\theta}_{j}^{(g)}\})$, achieves its maximum over all $M$ candidates grid $\{\gamma_{jm}^{(g)}\}_{1:M}$ over $\theta_j$'s domain $\Theta_j$.
\begin{equation}
\label{eq:profile}
m^{\star}_{jg} = \argmax_{m=1,\cdots,M} \sum_{k=1}^{K}
  f\left(\rt_{k}^{(g)},c = 1,\{\thetab^{(g)}_{-j},\, \theta_j {=} \gamma_{jm}^{(g)}\}\right), \qquad
\tilde{\theta}_j^{(g)} = \gamma_{jm^{\star}_{jg}}^{(g)}
\end{equation}
As a result, the recovery performance is valued by $\mathcal{R}(f) = \frac{1}{P} \sum_{j=1}^{P} \left( \tilde{\rho}_j - \tilde{e}_j \right) $, with $\tilde{\rho}_j = \corr_g(\tilde{\theta}_j^{(g)}, \theta_j^{(g)})$ and
$\tilde{e}_j$ the fraction of groups whose argmax sits on a grid edge. For detailed PySR setting and hyper-parameter selection, please check \autoref{app:SR}.

\subsection{Stages 3-4: screen and validate}
\label{sec:method-eval}
\paragraph{Screening criteria.} Each candidate expression $f \in \mathcal{F}^*$ is used
as the likelihood to fit every subject in the scoring bank with NUTS ($500$ warm-up and
$500$ sampling iterations, $2$ chains). For each parameter $\theta_j$, recovery is
measured by the correlation $r_j$ across subjects between the posterior mean
$\hat{\theta}_j^{(s)}$ and the generating value $\theta_j^{(s)}$. A parameter passes if
$r_j > r_0 = 0.5$ and fewer than half of its estimates lie within $1\%$ of the range from
either bound, since estimates that pile up at a bound can still correlate with the
truth. An expression passes if every parameter does. Convergence is monitored through
the per-subject $\widehat{R}$ and the number of divergent transitions, and the passing
expressions, which already trade density fit against complexity on the Pareto front, are ranked by their weakest parameter recovery, $\min_j r_j$.

\paragraph{Additional diagnostics.} Parameter recovery alone does not show that an
expression describes individual subjects well. We therefore also checked the validated
expressions with diagnostics that are standard in applied cognitive modeling
\citep{schad2021toward, gelman2020bayesian}. In addition to the convergence checks above ($\widehat{R}$ and divergent transitions), we ran posterior predictive checks and simulation-based calibration
\citep[SBC;][]{talts2018validating}. More specifically, the former metric compares choices and RTs simulated from each subject's posterior with the observed data, while the latter tests whether the sampler and the compiled
expression together return calibrated posteriors on data generated from the expression itself. Details and results are in \autoref{app:workflow}.

\section{Related Work}
\label{sec:related}

Recent symbolic distillation frameworks treat SR as the final stage of a surrogate-modeling pipeline, distilling the input--output behavior of a trained neural network, or of its components, into closed-form expressions
\citep{cranmer2020discovering,tan2026symtorch}. These frameworks target general input--output maps and impose no function-specific constraints before or during the search. Comparatively little work addresses probability densities specifically. Algorithms like MESSY \citep{tohme2024messy} guarantees a valid density by restricting the estimate to the maximum-entropy family, and \citet{liu2026symbolic} recover closed-form probability mass functions for discrete variables. The closest continuous analogue of the
surrogate-then-distill design, AI-Kolmogorov \citep{rajendram2026symbolic}, distills a nonparametric kernel density estimate with PySR under the MSE loss of
\autoref{eq: loss_mse}, so its result inherits the surrogate's error and does not depend on the model parameters $\bm{\theta}$. Our method differs in two ways. It checks that the neural likelihood preserves the parameters before distillation, so the resulting
expression is a function of $\bm{\theta}$ that can be used for inference, and it distills with an objective aimed at parameter recovery rather than density fit.

\section{Results}
\label{sec:results}

\subsection{Symbolic likelihoods recover the DDM parameters}
\label{sec:results-discovered}
\begin{figure}[!htbp]
\centering
\includegraphics[width=\linewidth]{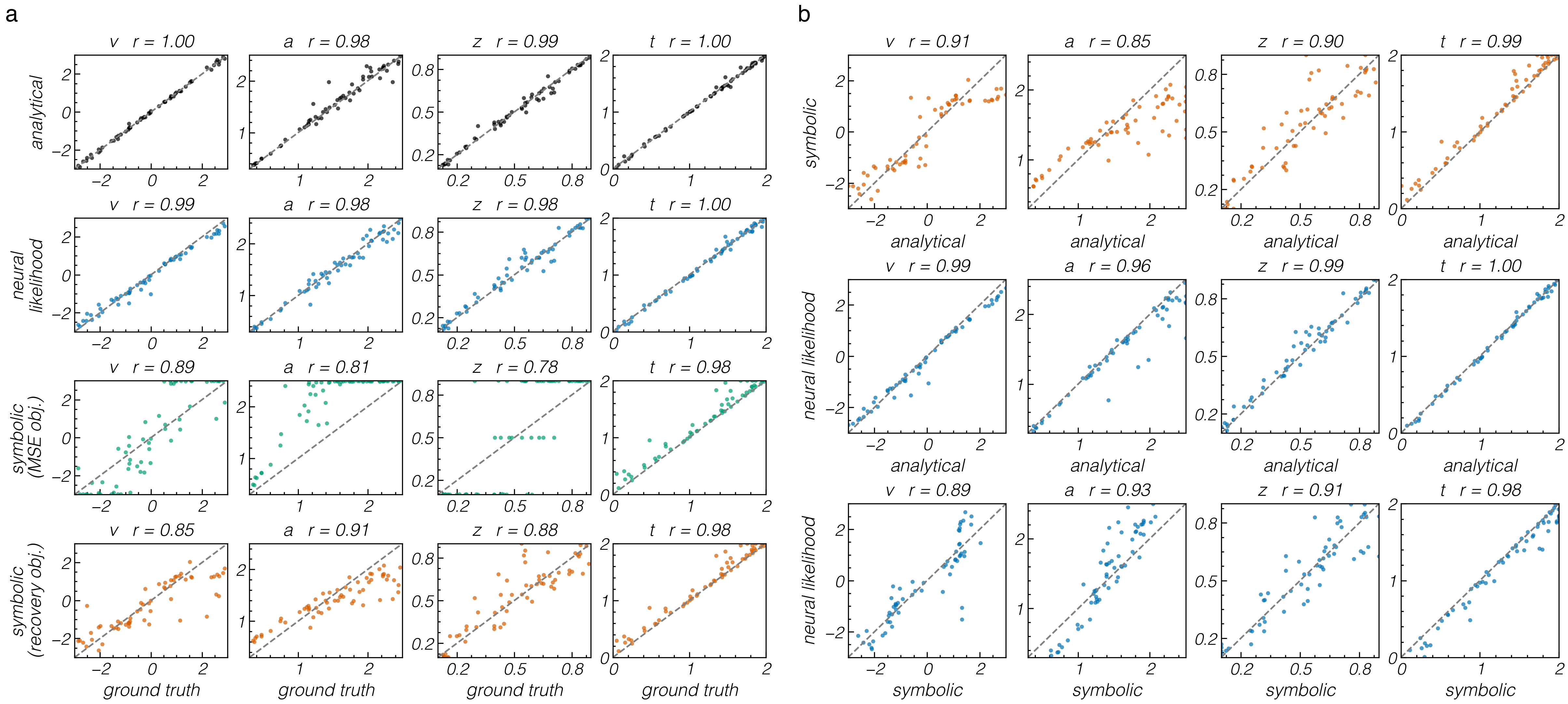}
\caption{\textbf{Parameter estimates on the test dataset.} Each point is one subject,
and columns are the four DDM parameters ($v$, $a$, $z$, $t$). Each panel title gives the
Pearson $r$ across subjects. \textbf{(a)}~True parameter value ($x$-axis) against the
posterior mean estimated with NUTS ($y$-axis) under four likelihoods, one per row. From
top to bottom, these are the analytical likelihood, the neural likelihood, an expression
found under Equation~\ref{eq: loss_mse}, and an expression found under Equation~\ref{eq:loss}
(third in \autoref{tab:top5}, fit in a separate NUTS run). \textbf{(b)}~Pairwise comparison of posterior-mode
estimates under the analytical likelihood, the neural likelihood, and the expression in
the bottom row of \textbf{a}.}
\label{fig:truth}
\end{figure}
We first asked whether the pipeline can produce a closed-form likelihood that recovers the parameters of the DDM. To answer this, we used $60$ simulated subjects with $1000$ trials each, none of whom were used to train the neural likelihood or to run the search (\autoref{app:data_split}). Recovery was measured for each parameter as the correlation across subjects between the true value and the posterior mean.
As a reference, we fit these subjects with the exact analytical likelihood, which shows how well the parameters can be recovered from this amount of data when the likelihood is known exactly. As expected, recovery was nearly perfect ($r = .98$--$1.00$ for all four parameters; \autoref{fig:truth}a, top row).

We then asked whether information was lost when the analytical likelihood was replaced by the neural likelihood in Stage~1. It was not. The neural likelihood recovered the parameters as well as the analytical likelihood did ($r = .98$--$1.00$; \autoref{fig:truth}a, second row), and the two gave nearly the same estimate for each subject ($r = .96$--$1.00$; \autoref{fig:truth}b, middle row). Because the expressions are distilled from the neural likelihood, not the analytical one, any further loss in recovery can be attributed to the distillation step (\autoref{app:nle}).

Finally, we fit the same subjects with a symbolic expression found under the
recovery-directed objective (\autoref{eq:loss}, $\lambda = 1$; third in
\autoref{tab:top5}). Recovery was lower than under the analytical likelihood but well above our criterion of $r > .5$ for every parameter ($r = .85$, $.91$, $.88$ and $.98$ for $v$, $a$, $z$ and $t$; \autoref{fig:truth}a, bottom row). 
To check whether the fixed search budget (\autoref{app:SR}) was enough to approximate the neural likelihood well,
we compared the expression's estimates with those of the neural likelihood. We found they agreed closely for each subject ($r = .89$--$.98$; \autoref{fig:truth}b, bottom row).

For comparison, the third row of \autoref{fig:truth}a shows an expression found under the standard MSE objective (\autoref{eq: loss_mse}). Its correlations looked nearly as good ($r = .89$, $.81$, $.78$, $.98$), but it failed our criterion because its estimates piled up at the bounds of the parameter range, for $a$ in $53\%$ of subjects and for $z$ in $85\%$. The correlations stayed high because whichever bound an estimate landed on still tracked the true value, which is why our criterion also rules out such estimates (\autoref{sec:method-eval}).

In addition to recovering the parameters, the distilled expressions are short enough to read. We next asked how many expressions in the full pool met our criterion. The pool contained $3371$ expressions from $124$ independent searches run under five values of the recovery weight $\lambda$ (\autoref{sec:results-objective}), and $276$ of them recovered all four parameters.
We ranked them by their weakest parameter, that is, by the lowest of their four correlations.
\autoref{tab:top5} lists the top five, and all five are written out in \autoref{app:forms}. Because the same $60$ subjects were used to rank the expressions, we also refit the top five on $90$ further subjects that were never used for any selection. All five still recovered every parameter (weakest $r = .81$ to $.90$; \autoref{tab:top5}).

\begin{table}[!htbp]
\centering
\caption{\textbf{The top five expressions' performance on the test bank.} $\lambda$ is
the recovery weight in \autoref{eq:loss}, $C_f$ is the expression complexity, MSE is its squared error against the neural log-likelihood on held-out rows (pool median $0.52$), and the
$v$, $a$, $z$ and $t$ columns give the recovery correlation $r$ for each parameter. Held-out gives the weakest of the four correlations on $90$ further subjects never used for ranking (\autoref{app:holdout}). KS is the median Kolmogorov--Smirnov distance between observed and posterior-predicted RTs (signed by choice), where lower
is better. For reference, it is $.287$ for the MSE expression in \autoref{fig:truth} and
about $.04$ for data simulated from the true parameters. A check mark means the
expression's KS is lower than that of an unbiased estimator with the same recovery
(\autoref{app:results-ppc}).}
\label{tab:top5}
\small
\begin{tabular}{@{}lrrrrrrrrrc@{}}
\toprule
 & $\lambda$ & $C_f$ & MSE & $v$ & $a$ & $z$ & $t$ & held-out & KS & beats null \\
\midrule
1 & $3$ & $44$ & $0.32$ & $0.96$ & $0.91$ & $0.89$ & $0.98$ & $0.87$ & $0.134$ & \checkmark \\
2 & $3$ & $25$ & $1.05$ & $0.94$ & $0.91$ & $0.88$ & $0.97$ & $0.90$ & $0.130$ & \checkmark \\
3 & $1$ & $42$ & $0.27$ & $0.88$ & $0.90$ & $0.88$ & $0.98$ & $0.81$ & $0.140$ & \checkmark \\
4 & $1$ & $45$ & $0.27$ & $0.88$ & $0.92$ & $0.88$ & $0.98$ & $0.84$ & $0.193$ & -- \\
5 & $3$ & $42$ & $0.51$ & $0.95$ & $0.88$ & $0.88$ & $0.98$ & $0.86$ & $0.160$ & \checkmark \\
\bottomrule
\end{tabular}
\end{table}

We highlight the second of these (Expression~2). Its weakest correlation is within $.004$ of the top-ranked expression, yet it is about half the size ($25$ nodes against $44$), and it had the best posterior predictive fit of the five ($\mathrm{KS} = .130$). The expression is given as follows:
\begin{equation}
\label{eq:expr387}
f = \frac{1}{a}\left[\,a \;-\; (\rt - t)\left(\Bigl(\frac{a^{2}e^{-2z}}{\rt - t} - v\Bigr)^{2}
    + e^{z^{2}}\right)\right].
\end{equation}
It recovered $v$, $a$, $z$ and $t$ at $r = .94$, $.91$, $.88$ and $.97$, with few estimates at the bounds (at most $10\%$ of subjects, for $a$).

The expression is also easy to interpret. Writing $\tau = \rt - t$ for the decision time and $D = a^{2}e^{-2z}$, its main term is $-\tau\,(D/\tau - v)^{2}/a$. 
This has the same form as the exponent of the Wald density, $-\tau\,(d/\tau - v)^{2}/2$, which is the leading term of the exact DDM first-passage density at short decision times, with $d$ the distance from the start point to the boundary \citep{navarro2009fast}. 
In both, the term is largest when the drift rate $v$ equals $d/\tau$, the rate at which evidence would have to accumulate to reach the boundary at exactly that time. 
The logarithm of that leading term, fit the same way, recovered the parameters about as well (weakest $r = .86$ on the held-out subjects, against $.90$ for Expression~2; \autoref{tab:holdout}), so the search found an expression as good as the known approximation without being given it.
Posterior predictive checks and simulation-based calibration for this expression are reported in \autoref{app:workflow} and \autoref{app:results-ppc}.

\subsection{The recovery-directed objective makes recovery possible}
\label{sec:results-objective}

\begin{figure}[!htbp]
\centering
\includegraphics[width=\linewidth]{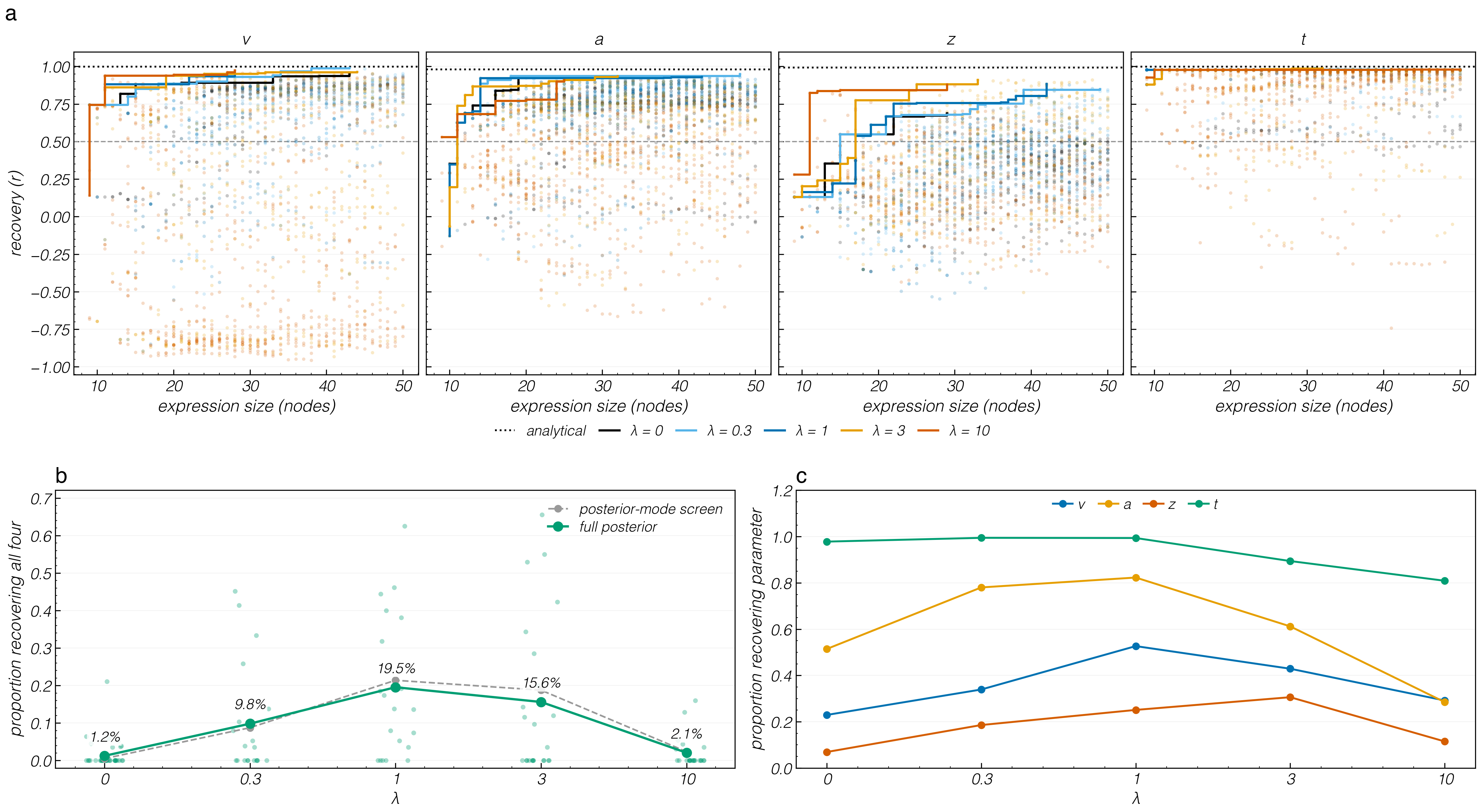}
\caption{\textbf{Effect of the recovery weight $\lambda$.} \textbf{(a)}~Recovery of each parameter against expression size. Each dot is one expression, colored by $\lambda$, and each line traces, for one value of $\lambda$, the best recovery reached at that size or smaller. 
The dotted line marks the analytical likelihood, and the dashed line the criterion of $r = .5$. \textbf{(b)}~Proportion of expressions that recovered all four parameters. 
Each translucent dot is one independently seeded search, and the solid line gives the pooled rate. The dashed line gives the faster posterior-mode (MAP) screen. \textbf{(c)}~Proportion of expressions that recovered each parameter. 
}
\label{fig:lambda}
\end{figure}

We next asked what allowed the search to find these expressions. Every expression discussed so far was found under a loss that included the recovery term. To isolate its effect, we compared searches that differed in the weight on that term, $\lambda$
(\autoref{eq:loss}), holding other settings the same (\autoref{tab:lambda}).
Without the recovery term ($\lambda = 0$, which reduces to plain MSE), $36$ independent searches produced $968$ expressions, and only $12$ of them ($1.2\%$) recovered all four parameters. The
drift rate and the start point were the main obstacles. Only $22.9\%$ of these expressions recovered $v$, and only $6.9\%$ recovered $z$. With the recovery term at $\lambda = 1$, $95$ of $486$ expressions ($19.5\%$) recovered all four parameters, and
the pass rates for $v$ and $z$ rose to $52.7\%$ and $25.1\%$ (\autoref{fig:lambda}b--c). Averaged over searches rather than expressions, the rates were $1.5\%$ and $18.5\%$, with non-overlapping $95\%$ intervals (\autoref{tab:lambda}).
Notably, without the recovery term, no expression recovered $z$ better than $r = .69$ (among those whose estimates did not pile up at the bounds), whereas every set of searches that included it exceeded $.85$ (\autoref{fig:lambda}a).

However, prioritizing recovery did not always help. Across the five values of
$\lambda$, the proportion of expressions that recovered all four parameters rose from $1.2\%$ at $\lambda = 0$ to $19.5\%$ at $\lambda = 1$ and then fell to $2.1\%$ at $\lambda = 10$ (\autoref{fig:lambda}b). The best weight also differed by parameter. Recovery of $t$ fell from $99\%$ at $\lambda = 1$ to $81\%$ at $\lambda = 10$, whereas $z$ was recovered most often at $\lambda = 3$ ($30.6\%$; \autoref{fig:lambda}c).

\subsection{Symbolic likelihoods are fast}
\label{sec:results-speed}

\begin{figure}[!htbp]
\centering
\includegraphics[width=0.8\linewidth]{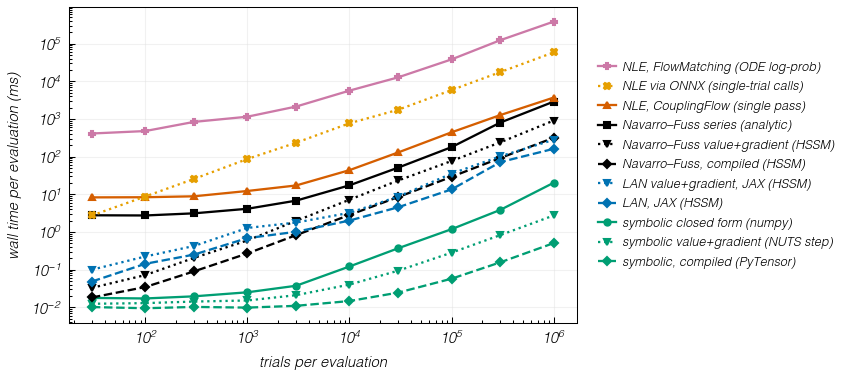}
\caption{\textbf{Evaluation cost across dataset sizes}, on log--log axes. Green curves are
Expression~2, black curves the analytical series and blue curves HSSM's LAN. Solid curves
are NumPy, dashed curves are compiled, and dotted curves include the gradient. The
remaining curves are neural likelihoods trained here, including a coupling flow that needs
a single network pass per evaluation. \autoref{tab:speed} gives the values at three sizes.}
\label{fig:speedcurves}
\end{figure}

We next asked how much the closed form saves in computation. We timed one summed log-likelihood for Expression~2, the analytical Navarro--Fuss series, a LAN \citep{fengler2021lans} and our neural likelihood, on datasets from $30$ to one million trials (\autoref{tab:speed}; \autoref{fig:speedcurves}). To make the comparison fair, the analytical series and the LAN were taken from HSSM and compiled, and each likelihood was timed alone and with the gradient that samplers like NUTS need at every step. At $1000$ trials, the size of one simulated subject in our tests, Expression~2 took $0.010$\,ms, or $0.015$\,ms with its gradient. This is $27$ to $39$ times faster than the compiled analytical series, $70$ to $84$ times faster than the LAN, and more than $100{,}000$ times faster than our neural likelihood, whose cost is dominated by integrating an ordinary differential equation (ODE). The advantage grew with the size of the dataset. At one million trials, Expression~2 took $0.51$\,ms against $315$\,ms for the compiled analytical series and $161$\,ms for the LAN.

\begin{table}[!htbp]
\centering
\caption{\textbf{Time to evaluate one summed log-likelihood} (ms; median of repeated
calls after warm-up). The expression example in \autoref{eq:expr387} was compiled through PyTensor, the analytical Navarro--Fuss series and the LAN are HSSM's compiled implementations (PyTensor and JAX), and the neural likelihood was timed at the $20$ ODE steps used in estimation. \emph{Gradient} is the value together with its gradient, which is what NUTS evaluates at each step.}
\label{tab:speed}
\small
\begin{tabular}{@{}lrrrrrrr@{}}
\toprule
 & \multicolumn{2}{c}{Expression~2} & \multicolumn{2}{c}{Navarro--Fuss} & \multicolumn{2}{c}{LAN} & neural \\
\cmidrule(lr){2-3}\cmidrule(lr){4-5}\cmidrule(lr){6-7}
trials & value & gradient & value & gradient & value & gradient & likelihood \\
\midrule
$1000$ & 0.010 & 0.015 & 0.27 & 0.59 & 0.68 & 1.3 & 1144 \\
$100{,}000$ & 0.057 & 0.28 & 29 & 76 & 13 & 35 & 38{,}080 \\
$1{,}000{,}000$ & 0.51 & 2.7 & 315 & 905 & 161 & 279 & 383{,}500 \\
\bottomrule
\end{tabular}
\end{table}

Expression~2 is fast because it needs neither a network nor a solver, nor an infinite series summed to a set precision. Every exponential in it depends only on the parameters, so the work for each trial is plain arithmetic. The price of this speed is the modest loss in recovery described in \autoref{sec:results-discovered}.

\subsection{Application to empirical data}
\label{sec:results-empirical}

So far, recovery and speed were both measured on simulated data. We therefore asked whether the symbolic likelihoods also hold up on real data. We refit a published dataset \citep[Cavanagh's frontal-theta study;][$3988$ trials, $14$ participants]{cavanagh2011subthalamic} as a hierarchical DDM in which every parameter varies by participant and drift rate also depends on frontal theta, as in the original analysis (\autoref{app:cavanagh}). We fit this model once with HSSM's compiled analytical likelihood and once with each of the $100$ best expressions, under identical priors and sampler settings. $96$ of the expressions converged ($\widehat{R} < 1.01$).

\begin{figure}[!htbp]
\centering
\includegraphics[width=\linewidth]{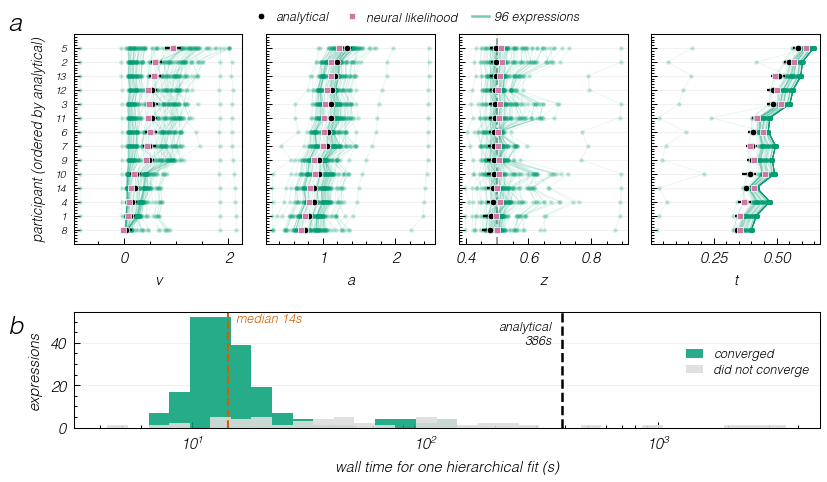}
\caption{\textbf{Hierarchical DDM fit to Cavanagh's data.} \textbf{(a)}~Participant-level posterior means in the full model, in which every parameter varies by participant. Each
translucent line is one of the $96$ converged expressions. Black points and bars show
the analytical likelihood's posterior mean and $89\%$ interval, and squares show the
neural likelihood's posterior mean. Within each panel, participants are ordered by that
parameter's analytical posterior mean, so rows do not correspond across panels. The
dashed line marks $z = 0.5$. \textbf{(b)}~Wall time for one fit of a simpler model, in
which only the drift rate varies by participant, under each of $267$ expressions ($213$
converged). Dashed lines mark the median expression ($14$\,s) and the analytical
likelihood ($386$\,s). All fits in \textbf{b} used four chains on one core of the same
machine.}
\label{fig:cavspeed}
\end{figure}

At the level of individual participants, the expressions ordered participants in nearly
the same way as the analytical likelihood on boundary separation $a$ (median rank
correlation $.97$), drift rate $v$ ($.93$) and non-decision time $t$ ($.90$;
\autoref{fig:cavspeed}a). They ranked participants poorly on $z$ ($.29$), mainly because participants barely differ on $z$ in these data (analytical posterior means from $.48$ to $.50$), and even the neural likelihood agreed with the analytical ranking at only $.72$.

Crucially, the symbolic likelihoods were also faster on these data. In the full model, the converged expressions took a median of $73$\,s against $174$\,s for the analytical likelihood. 
To time a larger set of expressions, we also fit a simpler model in which
only the drift rate varies by participant (\autoref{app:cavanagh}). Across $267$
expressions, the median fit took $14$\,s against $386$\,s, a $27$-fold speed-up, and
fewer than $2\%$ were slower (\autoref{fig:cavspeed}b). Both likelihoods needed about the same number of gradient evaluations, so the gain reflects the cheaper evaluation of each (\autoref{app:cavanagh}).

\section{Limitations \& Future Discussion}\label{sec:limitations}
A number of limitations warrant consideration. First, the discovered expressions are estimating functions rather than normalized densities, so they define only a generalized posterior \citep{bissiri2016general} whose intervals carry no calibration guarantee. Second, they estimate $t$ largely from the fastest responses, which makes them sensitive to fast guesses (\autoref{app:ddm-supp}). Third, every result here concerns the DDM, a deliberate test case whose exact likelihood let us check every step. The approach matters most for models without a tractable likelihood, such as those with collapsing bounds, and applying the recovery-directed objective to such models is the most important next step. Finally, the search produces hundreds of expressions that recover the parameters. Interpreting them as a set, for example by removing terms one at a time, could reveal which structures carry recovery.

\section{Conclusion}
\label{sec:conclusion}

In this study, we proposed a pipeline that distills a neural likelihood into a closed-form expression and asked whether such an expression can recover the parameters of a cognitive model. Using the DDM as a proof of principle, we found that it can and that this depends in large part on incorporating the recovery-directed objective. The resulting expressions are fast to evaluate, and an application to real data showed that this speed advantage carries over to hierarchical fits.

More broadly, our results suggest that the flexibility of simulation-based inference and the transparency of closed-form likelihoods need not be traded against each other. By asking a symbolic search to preserve properties important to the application (parameter recovery), it may be possible to have both.
\subsubsection*{Reproducibility Statement}
Every number in this paper is read from a committed result file produced by a single
open-source pipeline, so no claim rests on an unrecorded run. \autoref{sec:method} states
every setting at the level needed to judge the method, and the released repository records
the remainder needed to rerun it, including package pins, environment constraints, seeds and the
stage-by-stage runbook. All experiments run on CPU. The evaluation populations are
committed as data files with SHA-256 manifests because the simulator is not
seed-deterministic end to end; scoring identical subjects is what makes numbers comparable
across machines. The discovered expression discussed in the text is given in closed form
(\autoref{eq:expr387}), so the estimator can be used without re-running any search. Source code, trained weights, committed banks, and result files will be released publicly.

\bibliography{sections/references}
\bibliographystyle{iclr2027_conference}

\clearpage
\appendix
\section{SR implementation}
\label{app:SR}
The setting of PySR contains maximum complexity $50$ (node count), argument-complexity
caps of $9$ on the transcendentals, nesting constraints that forbid $\log \log$,
$\exp \exp$, and $\log \exp$, and $\max(31, 3 \times \text{threads})$ populations.
Everything else stays at PySR defaults. Runs are budgeted by wall clock ($1200$\,s
each).

\subsection{The objective-weight sweep}
\label{app:lambda}

The recovery weight $\lambda$ of \autoref{eq:loss} was swept over five values with the
operators, no-constants rule, search budget, test subjects and estimator held fixed.
Each value pools $20$ or more independently seeded searches, because the search, not the
expression, is the unit of replication. All expressions from one search share that
search, and the proportion recovering all four parameters varies from zero to over
$60\%$ across searches at the same $\lambda$. Averaged over searches, the rate is $1.5\%$ at $\lambda = 0$ and $18.5\%$ at $\lambda = 1$, with non-overlapping $95\%$ bootstrap intervals, and $15$ of the $20$ searches at $\lambda = 1$ found at least one expression that recovers all four parameters, against $8$ of $36$ at $\lambda = 0$. The effect of the recovery term therefore does not rest on a few searches. Three searches were run at a time
throughout, since a search run with a different number of threads explores differently.
The $\lambda = 0$ row pools two sets of MSE searches with identical settings (operators,
no-constants rule, $1200$\,s per search, maximum size $50$), $16$ searches on
ungrouped training rows and $20$ on the grouped training set shared with the other
values of $\lambda$. Both sets of rows were simulated and scored by the same neural
likelihood. The searches on the shared set recovered all four parameters less often
($3$ of $562$, $0.5\%$) than the others ($9$ of $421$, $2.1\%$), so pooling them, if
anything, understates the effect of the recovery term.

\begin{table}[!htbp]
\centering
\caption{Searches differing only in the recovery weight $\lambda$ of \autoref{eq:loss},
with identical operators, no-constants rule, neural likelihood, test subjects and
estimator (NUTS posterior means). Per-parameter columns give the proportion of
expressions that pass the criterion of \autoref{sec:method-eval} for that parameter.
\emph{Runs} is the number of independently seeded searches pooled at each value, and
pool sizes differ because each search ran for a fixed time. \emph{Any} is the number of those searches that found at least one expression recovering all four parameters. \emph{Pooled} gives that count and percentage over all expressions, and \emph{per search} gives the mean percentage over searches with a $95\%$ bootstrap interval. The $\lambda = 0$ row pools
two sets of MSE searches (see text). The last column is the median number of divergent
transitions per expression. Between $14\%$ and $41\%$ of expressions had no divergent
transitions, more so at the two ends of the sweep.}
\label{tab:lambda}
\small
\begin{tabular}{@{}lrrrrcrrrrr@{}}
\toprule
$\lambda$ & runs & any & pool & \multicolumn{2}{c}{recover all 4} & $v$ & $a$ & $z$ & $t$ & div. \\
 & & & & pooled & per search & & & & & \\
\midrule
$0$ & 36 & 8 & 968 & 12 (1.2\%) & 1.5 [0.5, 2.9] & 22.9\% & 51.4\% & 6.9\% & 97.8\% & 1 \\
$0.3$ & 20 & 10 & 560 & 55 (9.8\%) & 9.5 [3.7, 16.3] & 33.9\% & 78.0\% & 18.6\% & 99.5\% & 3 \\
$\mathbf{1}$ & 20 & \textbf{15} & \textbf{486} & \textbf{95 (19.5\%)} & \textbf{18.5} [10.9, 26.7] & \textbf{52.7\%} &
\textbf{82.3\%} & 25.1\% & \textbf{99.4\%} & 3 \\
$3$ & 24 & 12 & 634 & 99 (15.6\%) & 14.6 [7.0, 22.9] & 42.9\% & 61.2\% & \textbf{30.6\%} & 89.4\% & 7 \\
$10$ & 24 & 7 & 723 & 15 (2.1\%) & 2.1 [0.7, 4.0] & 29.2\% & 28.5\% & 11.5\% & 80.9\% & 4 \\
\bottomrule
\end{tabular}
\end{table}

\autoref{tab:lambda} gives the numbers behind \autoref{sec:results-objective}. The
proportion of expressions recovering all four parameters rises from $1.2\%$ at
$\lambda = 0$ to $19.5\%$ at $\lambda = 1$ and falls to $2.1\%$ at $\lambda = 10$, so the
recovery term helps only over a range of weights. With the density term negligible, the
search likely loses the signal that guides it toward sensible expressions, and the
recovery term alone is too coarse to replace it. Two further costs appear at high
$\lambda$. The non-decision time $t$, which is recovered under almost any structure,
drops from $99\%$ to $81\%$, so a search that stops caring about density fit loses the
parameter that was never at risk. The start point $z$, the hardest parameter at every
weight, is recovered most often at $\lambda = 3$ rather than at $\lambda = 1$, so the
weight that is best for all four parameters jointly is not the best for each one.
Divergent transitions are most frequent in the middle of the range, with a median of
$7$ per expression at $\lambda = 3$ against $1$ at $\lambda = 0$.

\section{The neural likelihood: architecture, gate, and checks}
\label{app:nle}

\paragraph{Training data hygiene.} Simulated trials enter the training bank only if they
are finite and $0 < \rt < 20$\,s (about $299{,}900$ of $300{,}000$ survive). The
classifier's output probabilities are clipped to $[10^{-6}, 1{-}10^{-6}]$ at evaluation.
All simulation goes through one entry point wrapping \texttt{ssm-simulators} at package
defaults, and the model registry checks its hard-coded parameter ranges against the
installed package, so that a breaking upstream change fails loudly instead of silently
shifting the prior.

\paragraph{Neural surrogate structure and training details.} The flow is a time-embedded
MLP of widths $[128, 128, 128]$ conditioned on $[\bm{\theta}, c]$, trained by conditional
flow matching to learn a velocity field that carries a base density to the target. The
classifier is a $2 \times 32$ $\tanh$ network with sigmoid output, trained with a
cross-entropy loss. The flow trains with Adam ($5 \times 10^{-4}$, batch $512$, $50$
epochs) and the classifier with Adam ($10^{-3}$, batch $1024$, $30$ epochs). The
classifier is trained first, because the flow-fitting library leaves torch in a no-grad
state. A post-training report checks for NaNs, under-training, and overfitting.

\begin{figure}[!htbp]
\centering
\includegraphics[width=\linewidth]{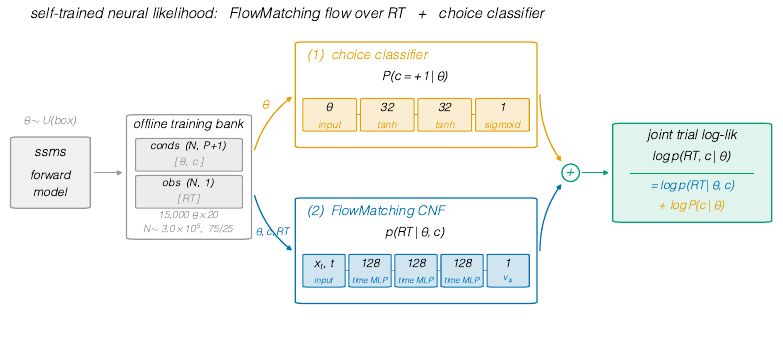}
\caption{\textbf{The neural likelihood.} The joint trial log-likelihood factorizes into
a flow over $\rt$ and a choice classifier (\autoref{eq:factorization}), trained on
$15{,}000$ parameter draws $\times\,20$ simulated trials. It is admitted only after
passing a diagnostic gate, with density calibration ECE $.041$ (gate $< 0.10$),
posterior-predictive KS median $.092$ (gate $< 0.15$), choice calibration MAE $.017$,
choice-rate agreement $r = .996$, and a cross-check against an independently
pre-trained LAN, $r = .982$.}
\label{fig:nle}
\end{figure}

\paragraph{Functional check.} Used as the likelihood under NUTS on the $60$ test
subjects, the gated neural likelihood recovers the parameters as well as the analytical
likelihood does ($r = .98$--$1.00$; \autoref{fig:truth}a, second row), so nothing
downstream is limited by the neural likelihood.

\section{Pipeline data split details}\label{app:data_split}

\paragraph{Stage 1.} All data come from committed banks of simulated subjects, each a
parameter set $\thetab \sim \mathrm{Unif}(\Theta)$ with up to $1000$ trials. The banks
are shipped as data files with SHA-256 manifests, because the simulator is not
seed-deterministic end to end and a re-simulated bank would therefore be different data.
The neural likelihood is trained on $15{,}000$ parameter sets with $20$ simulated trials
each (about $300{,}000$ trials), split 75/25 into training and validation.

\paragraph{Stage 2.} The symbolic regression training set scores simulated
upper-boundary trials with the neural likelihood, with rows $\x_i = (\thetab_i, \rt_i)$
and targets $\log p_{\phib}(\rt_i, c{=}1 \mid \thetab_i)$. It is organized as $G = 800$
parameter sets $\thetab^{(g)}$, $g = 1, \dots, G$, with $K = 20$ trials each ($16{,}000$
rows in total). The grouping is essential, because several $\rt$ draws at a fixed
$\thetab$ are what make the recovery proxy of \autoref{eq:profile} computable. The proxy
uses a precomputed block of $P \cdot G \cdot M \cdot K = 4 \cdot 50 \cdot 9 \cdot 5 =
9{,}000$ rows ($G = 50$ parameter sets, $M = 9$ grid values per parameter, and $K = 5$
trials), appended to the training array once.

\paragraph{Stages 3--4.} Two further disjoint banks of simulated subjects, with
$1000$ trials per subject, are used for scoring. Every candidate expression is scored
with NUTS on $60$ test subjects, which are used both to rank the expressions and to
report their recovery in \autoref{sec:results}. A further $90$ subjects from the same
bank were used by no step of the pipeline and serve only as the held-out check of
\autoref{app:holdout}. The faster posterior-mode screen of \autoref{app:screen} uses
$40$ separate subjects.

\section{The screen agrees on every marginal and disagrees on the winners}
\label{app:screen}

The posterior-mode screen and the full posterior (NUTS) agree closely on marginals. On
the $\lambda = 1$ searches, the mean number of parameters passed is $2.52$ against
$2.59$, the number passed per expression correlates at $.80$ between the two, and it is
identical for $71\%$ of expressions. That agreement is what makes a cheap screen look
trustworthy, until one asks which expressions it picks. The screen passes $104$
expressions and the full posterior passes $95$, but only $73$ are shared, so a
screen-gated pipeline would have promoted $31$ expressions the full posterior rejects
and discarded $22$ of the $95$ that recover. The screen also exaggerates the effect of
the objective. It puts the gain from $\lambda = 0$ to $\lambda = 1$ at about $41$-fold
($0.5\% \to 21.4\%$), where the full posterior puts it at about $16$-fold
($1.2\% \to 19.5\%$). The mode fit is the same kind of object as the recovery proxy the
search optimized, so part of the screen-level gain is the screen rewarding the search's
own shortcut. A cheap proxy that shares structure with the search loss cannot also be
the judge.

\section{The applied case: hierarchical DDM on Cavanagh}
\label{app:cavanagh}

\paragraph{Full model.} Every parameter varies by participant. Each of $v$, $a$, $z$ and
$t$ has its own group mean and spread, non-centred and mapped through a sigmoid onto the
simulator's range, so that draws stay in the region the expressions were validated on
without a clip that would remove the gradient. The upper bound on $t$ is set per
participant at that participant's fastest trial. Drift rate also carries the
frontal-theta covariate as a single population slope, as in the original analysis. The
$100$ expressions are the best of those that recover all four parameters, ranked by
their weakest correlation. Of these, $96$ converged ($\widehat{R} < 1.01$), at a median
of $73$\,s each, against $174$\,s for the analytical likelihood under the same model and
sampler settings.

Across the $96$, the between-participant spread relative to the analytical likelihood is
a median of $0.58$ for $v$ (inflated in $36\%$ of expressions), $0.81$ for $a$, $1.01$
for $t$ and $1.83$ for $z$, and rank agreement with the analytical likelihood is $.93$,
$.97$, $.90$ and $.29$, respectively. Thus $a$ and $t$ are ordered correctly at the
participant level, and $v$ is ordered correctly but compressed in spread. The
expressions also estimated non-decision time a median of $59$\,ms later than the
analytical likelihood did. Because the expressions are unnormalized, part of these group-level differences may come from the missing $\thetab$-dependent normalizer rather than from the fit to the data. The start point $z$ is ranked poorly (median $.29$, $10$th
percentile $-0.13$), but the participants barely differ on it. Their analytical
posterior means span only $.48$ to $.50$, and even the neural likelihood agrees with the
analytical ranking at only $.72$, so the ranking is dominated by small approximation
errors rather than by a failure specific to the closed forms.

\paragraph{Simpler model for timing.} To time a larger set of expressions under a common
model, we also fit a model in which only the drift rate varies by participant. Drift has
a population intercept ($\mathcal{N}(0, 1)$ prior), a frontal-theta slope
($\mathcal{N}(0, 0.5)$), and a non-centred participant offset whose spread has a
half-normal prior with scale $0.5$. The parameters $a \sim \mathrm{Unif}(0.3, 2.5)$,
$z \sim \mathrm{Unif}(0.1, 0.9)$ and $t \sim \mathrm{Unif}(0, t_{\max})$ are shared
across participants, with $t_{\max}$ just below the fastest response in the dataset. We
fit this model with the analytical likelihood and with each of $267$ expressions that
recover all four parameters, using $1000$ warm-up and $1000$ sampling iterations
in four chains on one core. Of the expressions, $213$ converged. The analytical
likelihood and a representative expression needed about the same number of gradient
evaluations ($108{,}392$ and $100{,}184$), so the speed-up reflects the cost of each
evaluation inside the fitted model ($3.7$ against $0.13$\,ms). Most expressions ($163$
of the $213$ that converged) placed the shared non-decision time at its upper bound, the
fastest response in the dataset. Among the $50$ that did not, the median fit was faster
still ($11$\,s), so the speed-up does not depend on these fits.

\section{Calibration and verification procedure}
\label{app:workflow}

Recovery across a population does not guarantee that an expression describes any
individual subject well. We therefore ran two further checks on the validated
expressions, each against a control that makes its number interpretable. The posterior
predictive check follows standard workflow tooling \citep{fengler2026hssm}. We fit
$\thetab$ with the expression as the likelihood, push each posterior draw through the
simulator, and compare the result with the observed trials by the KS distance on
choice-signed RTs. We judge it not against the predictive at the true $\thetab$, a
ceiling that no estimator with finite error attains, but against an unbiased estimator
perturbed to have the same recovery. Simulation-based calibration
\citep{talts2018validating} is run on data generated from the expression itself, with
$\exp f$ normalized on a fixed grid. It therefore tests the sampler, the compiled
expression and the normalization together, so any miscalibration on simulated data
would be attributable to the estimator rather than to this machinery. The analytical
likelihood is run through both checks unchanged as a control.

\section{What the discovered forms look like}\label{app:forms}

The five expressions of \autoref{tab:top5} are listed in full below. All use every
parameter, and none has estimates piling up at the bounds for any parameter (at most
$10\%$ of subjects). They come from two searches, with Expressions 1, 2 and 5 from one
search at $\lambda = 3$ and Expressions 3 and 4 from one at $\lambda = 1$, so they are
variations on a shared structure rather than five independent discoveries. All five
contain a term of the form $-\tau\,(D/\tau - v)^{2}$, up to a weight, with
$\tau = \rt - t$ and a different function $D$ of $a$ and $z$ that in every case grows
with $a$ and shrinks with $z$, so the interpretation given for Expression~2 in
\autoref{sec:results-discovered} applies to all of them. Expression~2 is the one printed
in \autoref{eq:expr387}, and Expression~3 is the one shown in \autoref{fig:truth}.

\begin{enumerate}\itemsep0.15em\small
\item $-a - z^{4} + z + \sqrt{a + |v|} + (t-\rt)\left[\left(\frac{a^{2}}{(t-\rt)(\sqrt{a}+16z^{4})} + v\right)^{2} + e^{z^{4}}\right]/a$
\item $\left[a + (t-\rt)\left(\left(\frac{a^{2}e^{-2z}}{t-\rt} + v\right)^{2} + e^{z^{2}}\right)\right]/a$
\item $-\frac{a^{2}}{a+v^{2}} + \sqrt{z} + z + (t-\rt)\left[z\left(v + \frac{e^{a}e^{-4z}}{t-\rt}\right)^{2} + \frac{e^{z^{2}}}{a^{3/2}}\right]$
\item $-\frac{a^{2}}{a+v^{2}} + \sqrt{z} + z + (t-\rt)\left[z\left(\frac{a^{2}e^{-4z^{2}}}{t-\rt} + v\right)^{2} + e^{2z^{4}}\right]$
\item $\sqrt{a} - a + 2\sqrt{z} - z^{2} + (t-\rt)\left[\left(\frac{a^{2}}{(t-\rt)(\sqrt{a}+16z^{4})} + v\right)^{2} + e^{z^{4}}\right]$
\end{enumerate}

\section{DDM setting details and supplementary analyses}\label{app:DDM}
\subsection{Setting details}
$\Theta = [-3,3] \times [0.3, 2.5] \times [0.1, 0.9] \times [0, 2]$ is the simulator's
own, and serves as the simulation range, prior support, and reference for the bounds
check throughout. Reflecting about zero gives an exact symmetry,
\begin{equation}
\label{eq:flip}
p(\rt, c{=}-1 \mid \thetab) = p(\rt, c{=}+1 \mid \flip(\thetab)), \qquad
\flip(v, a, z, t) = (-v,\, a,\, 1{-}z,\, t),
\end{equation}
so we model only the upper boundary and evaluate a lower-boundary trial at
$\flip(\thetab)$. The Navarro--Fuss series \citep{navarro2009fast} is used only as a
reference for measuring ceilings, never as a regression target, because the method is
meant for models that have no such formula.

\subsection{Supplementary analyses}
\label{app:ddm-supp}

\paragraph{Two choices in the recovery objective.} Both were validated against $422$
NUTS-scored expressions. The aggregation in \autoref{eq:loss} is a mean over parameters
rather than a minimum, because a minimum over four noisy estimates amplifies noise. And
the correlation term is kept even though the edge term alone scored slightly better on
the pools then available. An argmax away from the grid edge can be achieved by interior
optima that track nothing, and a search optimizing that statistic directly would hunt
for exactly that kind of shortcut.

\paragraph{Sampler settings do not change verdicts.} Refitting the two symbolic
expressions of \autoref{fig:truth}a on $30$ test subjects with $2000$ warm-up and
$2000$ sampling iterations in $4$ chains, instead of the standard $500$ and $500$ in
$2$ chains, moves recovery by at most $.04$ and never reverses which objective wins. The
recovery-objective expression goes from $.880/.872/.897/.984$ to $.917/.874/.901/.984$
for $v/a/z/t$, the MSE expression from $.855/.843/.722/.978$ to $.853/.844/.697/.978$,
and the median per-subject $\widehat{R}$ improves from $1.005$ to $1.001$. A broader
re-verification refit $473$ expressions drawn from all five values of $\lambda$,
selected as near the criterion ($267$), in the worst decile of divergent transitions
($157$), or in the worst decile of $\widehat{R}$ ($159$), and the groups overlap. Of
these, $12$ gained a pass, $3$ lost one, and $458$ were unchanged, with a median shift in
the weakest correlation of $-0.0001$ and $90\%$ of shifts within $[-.066, +.057]$.
Eleven of the $12$ gains were in the near-criterion group, so the standard settings were
not distorting verdicts.

\paragraph{Held-out subjects.}
\label{app:holdout}
The $60$ test subjects were used both to rank the expressions and to report their
recovery, so the top five could be favored by chance. The same simulated bank holds
$90$ further subjects that no step of the pipeline used. We refit the top five on these
subjects, together with the known leading-term form and the MSE expression of
\autoref{fig:truth}, using $2000$ warm-up and $2000$ sampling iterations in $4$ chains
(\autoref{tab:holdout}). All five expressions still recover every parameter. Their
weakest correlation falls by a median of $.02$ from the ranking subjects, the same as
for the known form, which was never selected, so most of the change reflects the new
subjects rather than the selection. The exceptions are Expressions 3 and 4, which fall
by $.07$ and $.03$ and come from the same search. The longer sampling moves no
correlation of the five by more than $.05$ relative to the standard settings, but it
does not resolve convergence for every subject. For Expressions 1, 3 and 4,
$\widehat{R}$ stays above $1.01$ for $25$ to $36$ of the $90$ subjects, whereas
Expression 5 converges for all of them and Expression 2 for all but $6$. Expression 2
has the most divergent transitions ($3303$, about $0.5\%$ of draws).

\begin{table}[!htbp]
\centering
\caption{Recovery on $90$ subjects never used for ranking, with $2000$ warm-up and
$2000$ sampling iterations in $4$ chains. \emph{Weakest} is the lowest of the four
correlations, and \emph{ranking} gives the same quantity on the $60$ subjects used to
rank the expressions (\autoref{tab:top5}). $\widehat{R} > 1.01$ counts subjects whose
chains did not converge, and \emph{div.} is the total number of divergent transitions.
The MSE expression fails the criterion because its estimates of $z$ sit at a bound for
$71\%$ of subjects.}
\label{tab:holdout}
\small
\begin{tabular}{@{}lrrrrrrrr@{}}
\toprule
 & $v$ & $a$ & $z$ & $t$ & weakest & ranking & $\widehat{R} > 1.01$ & div. \\
\midrule
Expression 1 & 0.91 & 0.90 & 0.87 & 0.99 & 0.87 & 0.89 & 25 & 28 \\
Expression 2 & 0.90 & 0.92 & 0.90 & 0.98 & 0.90 & 0.88 & 6 & 3303 \\
Expression 3 & 0.83 & 0.89 & 0.81 & 0.99 & 0.81 & 0.88 & 36 & 377 \\
Expression 4 & 0.84 & 0.90 & 0.86 & 0.99 & 0.84 & 0.88 & 33 & 12 \\
Expression 5 & 0.95 & 0.86 & 0.88 & 0.99 & 0.86 & 0.88 & 0 & 5 \\
Known leading-term form & 0.97 & 0.86 & 0.95 & 0.99 & 0.86 & 0.88 & 3 & 553 \\
MSE expression & 0.88 & 0.86 & 0.61 & 0.99 & -- & -- & 26 & 48 \\
\bottomrule
\end{tabular}
\end{table}

\begin{figure}[t]
\centering
\includegraphics[width=0.95\linewidth]{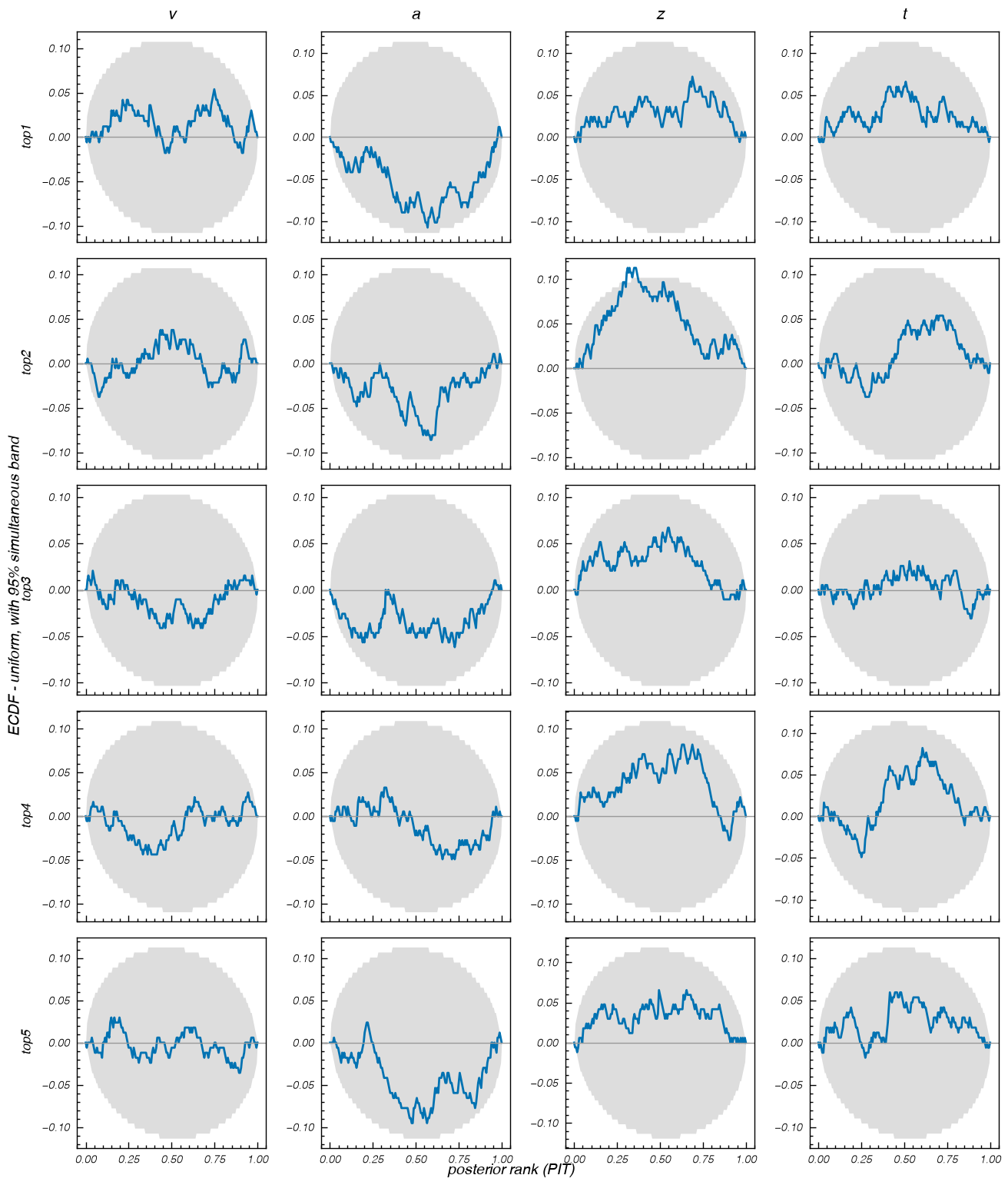}
\caption{Simulation-based calibration for all five expressions of \autoref{tab:top5}.
Each panel shows the difference between the empirical CDF of the posterior ranks and
the uniform CDF, with a $95\%$ simultaneous band \citep{saeilynoja2022graphical}, and a
calibrated posterior stays inside the band. Each row uses $200$ simulations of
$1000$ trials, with $\thetab$ drawn from the same uniform range as the recovery banks
and data generated from each expression's own grid-normalized density. No run produced
improper draws. Curves show the simulations with $\widehat{R} \le 1.05$ and bulk-ESS
$\ge 100$ ($167$--$194$ of $200$), and verdicts are unchanged on the full set. Nineteen
of the twenty parameter--expression combinations lie inside the band, the exception
being $z$ for Expression~2 (dispersion $1.03$, $\chi^{2}$ $p = .50$). The analytical
likelihood run through the same procedure is itself outside the band on $a$, which sets
the scale for reading a single excursion.}
\label{fig:sbc}
\end{figure}

\paragraph{Simulation-based calibration.} With $\exp f$ normalized on a fixed log-spaced
grid (grid and adaptive quadrature agree to within $1.6 \times 10^{-5}$ in $\log Z$) and
data generated from the expression itself, SBC rank statistics
\citep{talts2018validating} pass for all five expressions of \autoref{tab:top5}
(\autoref{fig:sbc}). Over $200$ simulations of $1000$ trials, rank dispersion for
$v/a/z/t$ is $1.02/1.01/0.95/1.01$, $0.98/0.98/1.04/0.92$, $1.02/0.97/1.06/1.01$,
$0.95/1.09/1.01/0.92$ and $1.03/1.00/0.96/0.99$ for Expressions 1--5, against
$0.92/0.90/1.01/0.95$ for the analytical likelihood run through the same procedure.
Nineteen of the twenty combinations sit inside the $95\%$ simultaneous band, and the
analytical likelihood is outside on one of its own four. The sampler, the compiled
expression and the normalization are therefore validated together. The scope of this
check should be stated precisely. Because the data are generated from the same
normalized density that serves as the likelihood, a failure would indict the machinery
rather than the model. The check does not cover the unnormalized $f$ used in the
pipeline, which differs from the normalized density by a $\thetab$-dependent term.

\paragraph{Do the recovered parameters reproduce behavior?}
\label{app:results-ppc}
\begin{figure}[!htbp]
\centering
\includegraphics[width=\linewidth,height=0.82\textheight,keepaspectratio]{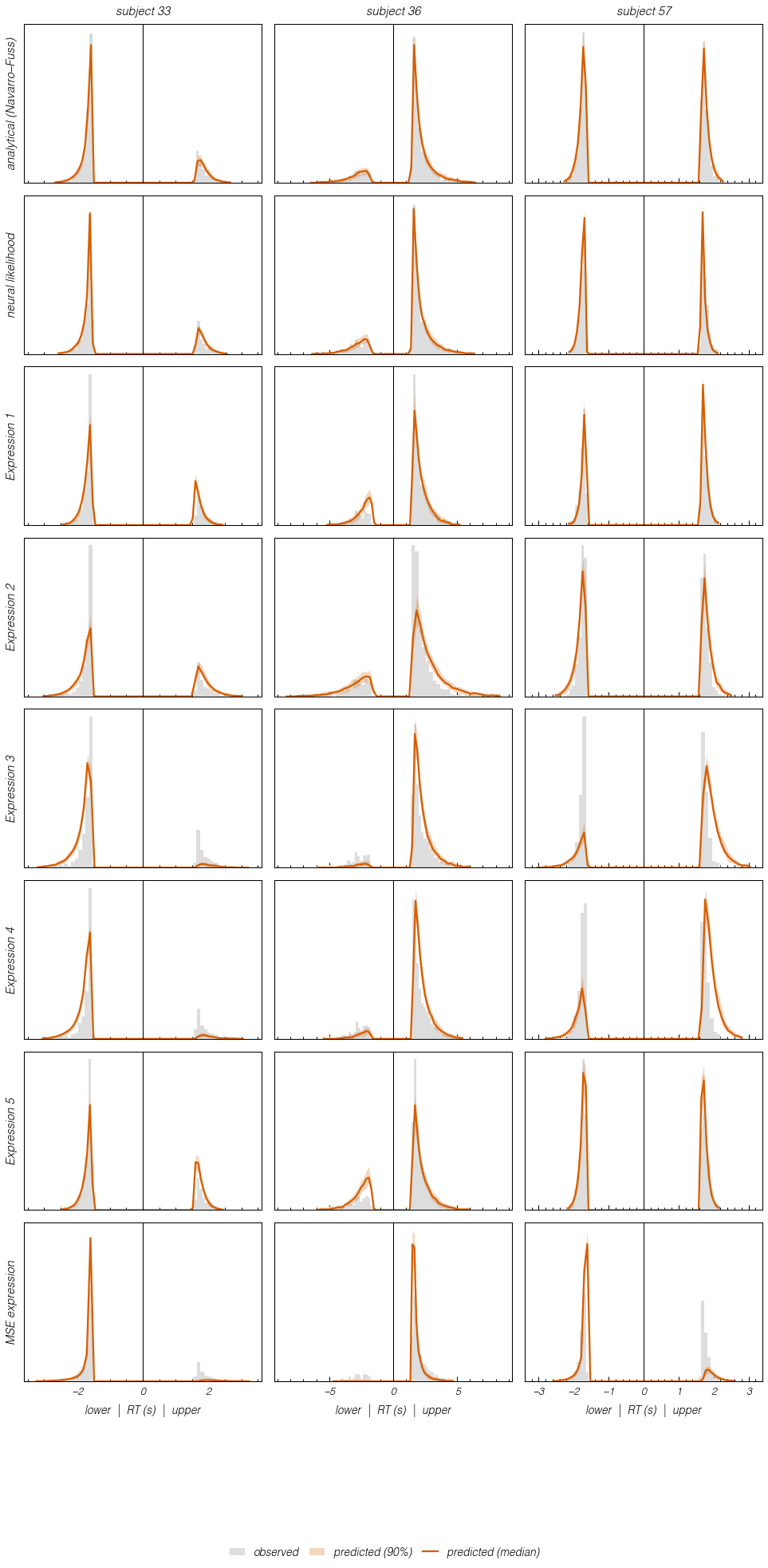}
\caption{Per-subject posterior predictive distributions for three subjects spanning the
observed choice balance. Each panel shows the observed choice-signed RT histogram
against predictive bands from one fitted likelihood. Rows are the analytical likelihood,
the neural likelihood, the five expressions of \autoref{tab:top5}, and the MSE
expression of \autoref{fig:truth}. The recovery-objective expressions track the RT
distributions at both boundaries, whereas the MSE expression visibly misallocates mass
(median KS values in \autoref{tab:top5}).}
\label{fig:ppc}
\end{figure}

Against the predictive at the true $\thetab$, the expressions fit worse (median KS $.130$
against $.036$ for Expression~2), but that reference is a ceiling no estimator with
finite error attains, so the comparison only restates that recovery is imperfect.
Against a fair null, an unbiased estimator perturbed to the expression's own recovery
$r$, Expression~2 does better ($.130$ against $.156$; \autoref{fig:ppc}), as do three of
the other four expressions, while the MSE expression reaches only $.287$ on the same
subjects. The predictive misfit of the recovery-objective expressions is therefore no
larger than their recovery level implies, and the objective improves predictive quality
as well as recovery.

\paragraph{Population-level estimator audit.} For each expression, we computed its
population argmax $\thetab^{*}(\thetab) = \argmax_{\vartheta} \E_{p(\cdot \mid \thetab)}
f(\vartheta, \cdot)$ by quadrature against the analytical density, which is the value
its estimates converge to with unlimited trials. This shows that $1000$ trials is
already effectively asymptotic. The recovery implied by $\thetab^{*}$ reproduces the
measured NUTS $r$ with median gaps of $0.000$ to $+0.022$ (correlations $.75$--$.95$ per
parameter), so the pass rates of \autoref{tab:lambda} are limited by the structure of
the expressions rather than by noise. The same audit locates the $z$ bottleneck. Among
the $487$ expressions scored at $\lambda = 1$ (the $486$ of \autoref{tab:lambda} plus a
reference closed form), the limit for $z$ clears the criterion for $141$ ($29\%$, close
to the measured $25.1\%$), against $399$ for $v$ and $479$ for $t$. The reference closed
form, the logarithm of the leading term of the small-time series
\citep{navarro2009fast}, is exponential-family separable in $u = \rt - t$, with
sufficient statistic $T(u) = (\sum u, \sum 1/u)$ and $\log u$ base measure, and $t$
enters only through the support. That role of $t$ is also the source of fragility.
Because $t$ is effectively estimated from the sample minimum, fast-guess contamination
(replacing trials, with $n$ fixed) reduces the number of the $96$ expressions at
$\lambda = 1$ that recover all four parameters (including the reference form) from $81$
to $53$, $29$, $2$ and $0$ at $0$, $0.25$, $0.5$, $1$ and $2\%$ contamination. This is a
breakdown point of order $1/n$, as expected for a parameter estimated from the sample
minimum.

\section{AI use statement}
\label{app:ai}
In this work, we used generative AI tools (Claude, Anthropic, through Claude Code) to
generate the initial scaffold of the pipeline code. We have not used generative AI tools
to propose the research question or the recovery-directed objective, or to interpret
results, and generating synthetic data sets, formulating mathematical claims, writing
proofs, translation and qualitative data analysis are not applicable to this work.
Additionally, we used generative AI tools to help write and organize the results and the appendix. We have reviewed all AI-assisted work. All code was verified by
the authors, and every number in the paper was checked against the result files the code
produced. We take responsibility for the final content of this work, including text,
claims or artifacts produced with the aid of generative AI.

\end{document}